\documentclass[journal,transmag]{IEEEtran}
\usepackage{graphicx}

\begin{document}
\title{A radial basis function neural network based approach for the electrical characteristics estimation of a photovoltaic module}

\author{\IEEEauthorblockN{Francesco Bonanno\IEEEauthorrefmark{1},
Giacomo Capizzi\IEEEauthorrefmark{1},
Christian Napoli\IEEEauthorrefmark{2,*}, 
Giorgio Graditi\IEEEauthorrefmark{3}, and
Giuseppe Marco Tina\IEEEauthorrefmark{1}}
\IEEEauthorblockA{\IEEEauthorrefmark{1}Dpt. of Electrical, Electronics and Informatics Engineering, University of Catania, Italy}
\IEEEauthorblockA{\IEEEauthorrefmark{2}Dpt. of Physics and Astronomy, University of Catania, Italy}
\IEEEauthorblockA{\IEEEauthorrefmark{3}Centro Ricerche ENEA, Portici (NA), Italy}
\thanks{*Email: napoli@dmi.unict.it.}
\thanks{PUBLISHED ON: \bf Applied Energy, vol. 97, pp. 956-961, 2012}}

\markboth{RBFNN approach for electrical characteristics of a PV module -- PREPRINT}%
{Shell \MakeLowercase{\textit{et al.}}: Bare Demo of IEEEtran.cls for Journals}

 \begin{titlepage}
 \begin{center}
 {\Large \sc PREPRINT VERSION\\}
  \vspace{15mm}
 {\Huge A radial basis function neural network based approach for the electrical characteristics estimation of a photovoltaic module\\}
 \vspace{10mm}
 {\Large F. Bonanno, G. Capizzi, C. Napoli, G. Graditi and G.M. Tina\\}
 \vspace{15mm}
 {\Large \sc PUBLISHED ON: \bf Applied Energy, vol. 97, pp. 956-961, 2012.\\}
 \end{center}
 \vspace{10mm}
 {\Large \sc BIBITEX: \\}
 \vspace{5mm}
 
@article\{Bonanno2012956,\\
author = "F. Bonanno and G. Capizzi and G. Graditi and C. Napoli and G.M. Tina",\\
title = "A radial basis function neural network based approach for the electrical characteristics estimation of a photovoltaic module ",\\
journal = "Applied Energy ",\\
volume = "97",\\\
number = "0",\\
pages = "956 - 961",\\
year = "2012",\\
issn = "0306-2619",\\
doi = "http://dx.doi.org/10.1016/j.apenergy.2011.12.085",\\
url = "http://www.sciencedirect.com/science/article/pii/S0306261911008919",\\
\}
 \vspace{35mm}
 \begin{center}
Published version copyright \copyright~2012 ELSEVIER \\
\vspace{5mm}
UPLOADED UNDER SELF-ARCHIVING POLICIES\\
NO COPYRIGHT INFRINGEMENT INTENDED \\
 \end{center}
\end{titlepage}

\IEEEtitleabstractindextext{
\begin{abstract}
The design process of photovoltaic (PV) modules can be greatly enhanced by using advanced and accurate models in order to predict accurately their electrical output behavior. The main aim of this paper is to investigate the application of an advanced neural network based model of a module to improve the accuracy of the predicted output I--V and P--V curves and to keep in account the change of all the parameters at different operating conditions. Radial basis function neural networks (RBFNN) are here utilized to predict the output characteristic of a commercial PV module, by reading only the data of solar irradiation and temperature. A lot of available experimental data were used for the training of the RBFNN, and a backpropagation algorithm was employed. Simulation and experimental validation is reported.
\end{abstract}

}

\maketitle

\section{Introduction}
\IEEEPARstart{R}{ecent} advances in solar cells manufacturing require dynamic modeling and characterization of the developed commercial PV module. Depending upon the application, several models of solar cells have been proposed in literature to consider various details of the systems powered by PV module as a whole. Several circuits with different topologies that model photocurrent sources, current leakage paths and loss elements are also provided. Several analytical modeling techniques based on complex mathematical expressions to estimate the behavior of the solar cells have been reported in literature. Several models for solar cell utilize non-linear lumped parameter equivalent circuits and their parameters are determined by experimental current--voltage characteristics using analytical or numerical extraction techniques \cite{a1}, \cite{a2} and \cite{a3}. In \cite{a4} the parameters extraction is performed starting by the consideration of the temperature and maximum power point (MPP) voltage and current distributions versus solar irradiance. Precisely the authors perform the identification of photovoltaic array model parameters by robust linear regression methods. The method allows to obtain the parameters by using only the solar irradiance. In \cite{a5} a neural network based approach for solar array modeling is based on non-linear radial basis function networks that are connected to the input vector in straightforward manner. The basic and some remarks about their training procedure are presented and applied to simulate the current and voltage curves of the PV array. In \cite{a6} RBFNN is are applied to model I--V curves and MPP of a PV module. The main issue is the use of a genetic algorithm-based RBFNN training schematic in order to obtain an optimal number of radial basis functions by using only input samples of a commercial PV module. In \cite{a7} the model proposed by the authors require five parameters accounting the solar irradiance and the PV module temperature. The authors state that is useful to predict the performance for engineering application. The key point should be its powerful management by the use of only the few data provided by manufacturers. By using soft computing as neural networks the input data have to be provided only at the first stage being the model robust at different operating conditions. A RBFNN as we propose here is able to generalize and work are in progress to do extraction of circuital parameters. This is only a refinement of the proposed RBFNN based model because at present time are available simple formulas that provide analytical expression for the generated electric current and power. In \cite{a8} the authors mainly discuss about the photovoltaic module and the relative experimental set-up to be used to validate a simplified four parameters model. The shunt resistor of a circuital model for PV id here neglected but the authors have the merits to calculate the electric power of a PV module by its simple formula. The five parameter model is an implicit non-linear equation, which can be solved with a numerical iterative method such as Levenberg--Marquardt algorithm. In \cite{a9} the authors investigate the thermal and electrical performance of a solar PV thermal collector. A more detailed coupled thermal and electrical model for PV module was developed in the past by the last author of the present paper. Then the relative work is beyond the scope of the present paper that mainly aim to the electrical characteristics estimation of a PV. This topic was subject of matter of other research paper. In \cite{a10} the paper deal with the photovoltaic field emulation including dynamic and partial shadow conditions and is basically based on the experimental set-up in the assembled PV plant. In the present paper the our trained RBFNN is used to obtain the current--voltage (I--V) characteristics and power--voltage (P--V) curves of a commercial PV module. At present time MPP predictions by using neural networks is not considered in the paper. The computation time is very low when compared with conventional numerical simulation methods \cite{a11}. The PV module I--V characteristics are now calculated in a novel matter and are related to the centroids of the selected RBFNN. Furthermore the circuital model parameters as the solar cell series resistance, the solar cell parallel resistance, the short circuit current at a fixed set point temperature and the ideality factor, can be calculated. The proposed calculation procedure is based on simple formulas and on the fundamentals of circuit theory then being the main advantage the use of the main parameters of the selected RBFNN. In \cite{a6} a good method is presented and described. To validate the proposed RBFNN model a set of solar radiation data was used as input. The data were collected by an experimental set up assembled in the laboratory of Power Systems at the Department of Electrical, Electronic and Informatics engineering of the University of Catania. The main I--V and P--V characteristics are now available and especially at high levels of solar radiation prediction is accurate. The solar cell is usually represented by a simple electrical circuit with series and shunt resistances and a diode. The I--V characteristics of the solar cell can be simulated using the electrical circuit parameters. The main solar cell parameters, e.g., the open circuit voltage ($V_{OC}$), short circuit current density ($I_{SC}$), fill factor (FF), external quantum efficiency (EQE), and conversion efficiency ($\eta$), can be obtained from the simulated I--V characteristics. Generally the researchers in this field start with a simple model of solar cells and add more elements to the circuit to make sophisticated and comprehensive models for system analysis used in detail-oriented applications. In the past the authors have gained experiences in the application of artificial and hybrid neural networks architectures in different research fields \cite{a12}. Finally in this paper, we proposed a novel neural network based technique for the modeling of complex solar cell characteristics. The paper is organized in five sections: in the first one, the introduction, a general description of the paper is presented, in the second, the conventional equivalent circuit of a solar cell model is reviewed. Variations of the PV equivalent circuit parameters under different operating conditions will be investigated. In the third, the structure of the selected neural network, the training procedures and testing are presented. In the fourth, the proposed model with the data available from the experimental set up is presented. In the last part conclusions are drawn.
\begin{figure*}[t]
\includegraphics[width=.95\textwidth]{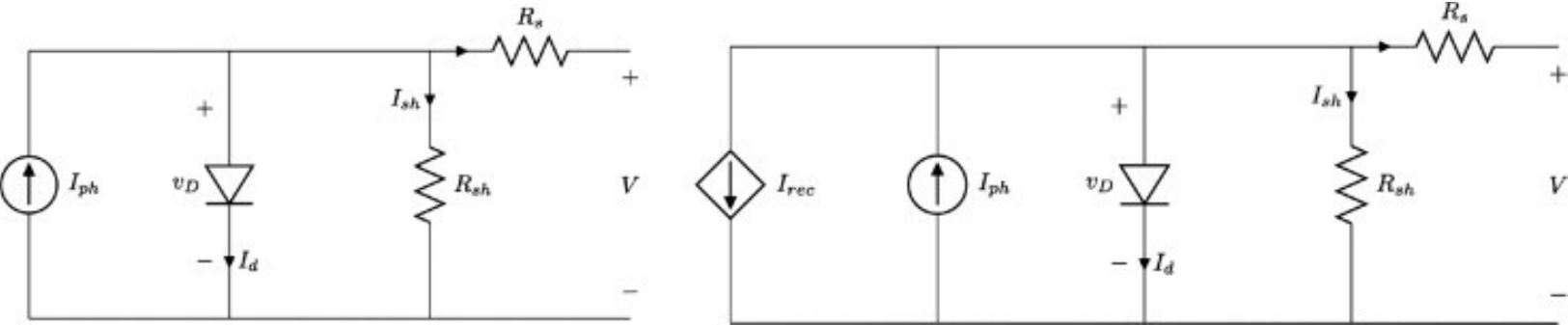}
\caption{Left: Equivalent circuit of PV solar cell used in the five parameters model. In the right figure with a current sink ($I_{rec}$) accounting the current losses due to recombination phenomena.}
\label{f1}
\end{figure*}

\section{Circuital models for PV modules}
Accurate extraction and optimization of photovoltaic (PV) cells and PV module parameters are very important to improve the device quality during fabrication and in device modeling and simulation, also for monitoring and diagnostic of PV array \cite{a1}. The most common model for a PV module is the five-parameter and a PV performance model that is derived from an equivalent circuit of a solar cell, which consists of a current source, a diode, and a series and parallel resistors, as shown in Fig. 1. Starting from a current balance at node A, substituting in Ohms law and the Shockley diode equation for the currents through the resistors and the diode, respectively, yields the model characteristic equation, given by the (\ref{e1}). The variable $I_0$ is the reverse-bias saturation current and a is the ideality factor, and $V_t$ is the thermal voltage defined by (\ref{e2}), where $N_s$ is the number of solar cells in series, $k$ is Boltzmanns constant, $T$ is the cell temperature and $q$ is the charge of an electron. 

\begin{equation}
I=I_{ph} - I_0 \left( e^\frac{V+IR_s}{R_{sh}} -1  \right) -\frac{V+IR_s}{R_{sh}}
\label{e1}
\end{equation}

\begin{equation}
V_t = \frac{N_s k T}{q}
\label{e2}
\end{equation}

The current source $I_{ph}$ represents the charge carrier generation in the semiconductor layer of the PV cell caused by the incident radiation. The shunt diode represents recombination of these charge carriers at a high forward-bias voltage. The parallel resistor $R_{sh}$ means high-current path through the semiconductor along mechanical defects and material dislocations. The series resistor ($R_s$) embodies series resistance in the outer semiconductor regions, primarily at the interface of the semiconductor and the metal contacts. The five-parameter equivalent circuit accurately predicts the performance of crystalline solar module under varied operating conditions but it does not perform as well as for amorphous and thin-film technologies. To improve the performances of this model, some modifications have been introduced in both the definitions of the parameters and the in the topology of the circuit reported in Fig.~\ref{f1}. A more complex model, called the seven-parameter model, based on the equivalent circuit model of a PV cell has been developed in \cite{a2}. This model is an extension of the six-parameter model, which is currently used by the California Commission (CEC) and is one of the models in the Solar Advisory Model (SAM) being developed by NREL. The two new parameters are respectively: the non-linear series resistance temperature dependence $\delta$ as given in (\ref{e3}) and the  diode reverse saturation current radiation dependence $m$ as in (\ref{e4}):

\begin{equation}
R_s(T)={R_s}_{{}_{\mbox{\tiny ref}}} e^{\delta(T-T_{{}_{\mbox{\tiny ref}}} )}
\label{e3}
\end{equation}

\begin{equation}
I_0(G,T)= {I_0}_{{}_{\mbox{\tiny ref}}} \left(\frac{G_{{}_{\mbox{\tiny ref}}}}{G}\right)^m \left(\frac{T}{T_{{}_{\mbox{\tiny ref}}}}\right)^3 e^{\frac{1}{k}\left[\left(\frac{E_g}{T}\right)_{{}_{\mbox{\tiny ref}}}-\left(\frac{E_g}{T}\right)\right]}
\label{e4}
\end{equation}
 
In amorphous silicon solar cells, the main part of the PV generation occurs in the intrinsic i-layer. As the solar cell ages, the degrading state creates a recombination center that sinks some of the generated current and reduces the power of the solar cell. A circuit proposed in \cite{a3}, that includes an additional current sink, shown in Fig.~\ref{f1}, seeks to point out the recombination current in the middle intrinsic layer of an amorphous silicon cell from the current in the outer semiconductor regions. Finally the output power of a PV module can be calculated as the product of terminal voltage and current by the following equation

\begin{equation}
P=I_{ph}V-I_0\left[e^{k\left(\frac{V}{N_s}+ IR_s\right)}-1\right]V-\left(\frac{V}{N_s}+IR_s\right)\frac{V}{R_{sh}}
\label{e5}
\end{equation}

\section{Model of solar cell: two-diode model and lumped resistances-single diode, parallel resistor and the five parameters model}
A more complex and accurate mathematical expression for a single solar cell is known as the double diode model. It is used to describe more accurately the current density--voltage (J--V) characteristics of such cells when the incident light intensity varies. The two-diode model is described in the work of Nann and Emery \cite{a4}, which is also an excellent source of references for prior work done regarding two-diode models of PV devices. The I--V characteristics obtained from this model are summarized by (\ref{e6}), which expresses the total measured current (I) as a function of the applied external voltage (V). The sign convention used in (\ref{e6}) is such that active power production occurs in the fourth quadrant of the I--V Cartesian coordinate plane.

\begin{equation}
I=I_{ph}V-I_{01}\left(e^{\frac{V+IR_s}{\eta_1 V_t}}-1\right)-I_{02}\left(e^{\frac{V+IR_s}{\eta_2 V_t}}-1\right)-\frac{V+IR_s}{R_{sh}}
\label{e6}
\end{equation}

The terms $I_{01}$ and $I_{02}$ represent the reverse saturation currents for each one of the two diodes. The first term represents the contribution from an ideal diode with $\eta_1$ diode--quality factor, whose value is usually taken as 1. The second term represents the contribution from a secondary diode with quality factor $\eta_2$, whose value is usually taken as 2. It is allowed to manifest values between 1 and 2. The term $I_{ph}$ is the photogenerated current, which is a function of irradiation. Normally this module is applied to c-Si cells, but it can be applied to modelize Thin-Film Photovoltaic Modules (e.g. GIGS \cite{a13}). In the ideal model the dc current generated from the current source is proportional to the solar irradiance. There are other parasitic sources in solar cells that affect the current generation and the terminal voltage. These parasitic behaviors can be modeled in an equivalent circuit as a series resistance to model the voltage drop and a current leaks proportional to the terminal voltage. The current leak is distributed throughout the p--n junctions and may not always be represented as a lumped element. The series and parallel resistances means a practical technique that can be used to consider these effects.

\section{PV module modeling using RBFNN}
\begin{figure*}[t]
\includegraphics[width=.95\textwidth]{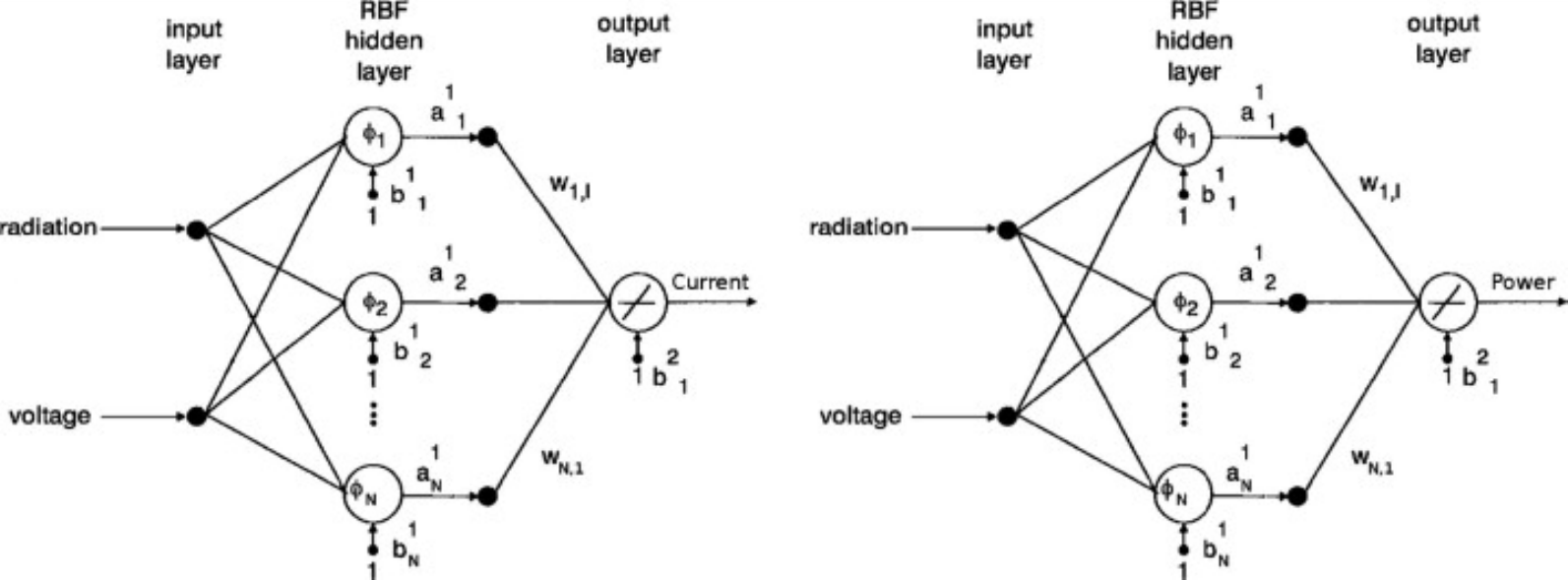}
\caption{ PV array RBF network structures.}
\label{f2}
\end{figure*}

The RBFNNs proposed to represent a PV array are shown in Fig.~\ref{f2}. These networks are composed by three layers: the input, hidden and output layers. The input layer consists of a three-dimensional vector, {\bf X}, whose elements are solar radiation, ambient temperature and load voltage. The output layer, {\bf Y}, has only one element, the load current, though in general it can be a vector of any dimension. The hidden layer is composed of L radial basis neurons, $\psi_j (1 \leq j \leq L)$, that are connected directly to all the elements in the input layer \cite{a6}. For a data set consisting of $N_T$ input vectors together with the corresponding output currents there are $N_T$ such hidden units, each corresponding to one data point. For $1\leq n \leq N_T$ follows:

\begin{equation}
\mathbf{X}^n = \left [
\begin{array}{l}
R^n_{ad}\\
T^n_d\\
V^n
\end{array}
\right] ~~~~~~~
\mathbf{Y}^n = \mathbf{I}^n
\label{e7}
\end{equation}
the hidden unit can be expressed as a matrix:
\begin{equation}
\mathbf{\Phi} = \left [
\begin{array}{cccc}
\phi^1_1 & \phi^1_2 & \ldots & \phi^1_L  \\
\phi^2_1 & \phi^2_2 & \ldots & \pi^2_L  \\
\vdots & \vdots & \ddots & \vdots \\
\phi^{N_T}_1 & \phi^{N_T}_2 & \ldots & \phi^{N_T}_L  \\

\end{array}
\right] 
\label{e8}
\end{equation}
and the weight vector
\begin{equation}
\mathbf{W} = \left [
\begin{array}{c}
W_{1,k}\\
W_{2,k}\\
\vdots \\
W_{L,k}
\end{array}
\right] 
\label{e9}
\end{equation}

The function $\phi_j(\mathbf{X}^n)$ takes the form of a non-linear distribution. The one commonly used is the Gaussian function of the form
\begin{equation}
\mathbf{\Phi}_j(\mathbf{X}^n) = e^\frac{| \mathbf{X}^n-\mathbf{\mu}_j |}{\sigma^2_j}
\label{e10}
\end{equation}
where $\mathbf{\mu}_j (1 \leq j \leq L)$  is a vector having the same dimension as $\mathbf{X}$ and representing the center of the RBF, $\mathbf{\Phi}_j$, and $\sigma$ is a scalar defining the width of an RBF, sometimes called the spread width. Radial basis networks can require more neurons than standard feedforward backpropagation networks, but often they can be designed in a fraction of the time it takes to train standard feedforward networks. They work best when many training vectors are available. According to the general principle of the RBF network mapping, the proposed RBFNN for a PV array is shown in Fig.~\ref{f2}.

\section{The proposed RBFNN based model}
The training of an RBFNNs network involves the determination of a number of RBFs and optimal values of the centers, weights and biases \cite{a14}, \cite{a15}, \cite{a16} and \cite{a17}. The criterion is to minimize the sum of squared errors. The RBFNN can be designed and trained in a smaller time with respect of a standard feed-forward neural network, even if they show a greater number of neurons. In a RBFNN the inputs are directly connected to the radial basis neuron, but in this particular case, they differ respect to the other neuron types. In fact the input comes from the evaluation of the biased norm relative to the difference between the input value and the data centroid which can be used as statistical weight for the input itself. The radial basis functions neural network (RBFNN) consists of an hidden radial basis layers and a an output linear layer, the transfer function for a radial basis neuron is a gaussian like finite $L^2$ function, typically of the kind of $exp (−n^2)$. The network is build up iteratively by adding a neuron at any time step, so on until the mean squared error falls into the expected limits \cite{a19}, \cite{a20} and \cite{a21}. We have utilized a 16 by one RBFNN with an hidden layer of 16 radial basis neurons and one output linear neuron, the network was trained with a random set of 5600 bidimensional input [G, V] and related target output [I(V, G)], where the solar irradiation G is ranging from 200 $Wm^{−2}$ to 1000 $Wm^{−2}$ under a voltage between 0 V and 30 V. The trained network was then tested for the prediction of the all data series providing a correct response with less than 2\% of relative mean squared error. By using the same configurations the network was also trained with a random set of 4600 bidimensional input [G, V] and a related target output [W(V, G)], for the same solar irradiation and voltage intervals. The trained network was tested again for the prediction of the data series as a whole. It perovides a correct response with less than 1\% of relative mean squared error. The type and the structure of the selected networks lead to the analytic expression of the electric current I (\ref{e11}) and of the Power P (\ref{e12}) as functions of the voltage V and irradiation G. The parameters involved in the Eqs. (\ref{e11}) and (\ref{e12}) are reported in Table~\ref{t1}.

\begin{equation}
I=f(V,G)\approx \sum\limits_{i=1}^{16} w_i^{(1)}e^{-\frac{\left(V-c_{i1}^{(1)}\right)^2\left(G-c_{i2}^{(1)}\right)^2}{\sigma^2}}
\label{e11}
\end{equation}

\begin{equation}
P=f(V,G)\approx \sum\limits_{i=1}^{16} w_i^{(2)}e^{-\frac{\left(V-c_{i1}^{(2)}\right)^2\left(G-c_{i2}^{(2)}\right)^2}{\sigma^2}}\label{e12}
\end{equation}

\begin{table}
\centering
\small
\caption{Weight and centroids of the implemented neural networks}
\label{t1}
\begin{tabular}{|r| l l l l l l|}
\hline
$i$ & $w_i^{(1)}$ & $c_{i1}^{(1)}$ & $c_{i2}^{(1)}$ & $w_i^{(2)}$ & $c_{i1}^{(2)}$ & $c_{i2}^{(2)}$\\
 \hline
1 & 5.46 & 12.56 & 200 & -508.62 & 13.05 & 200 \\
2 & 3.40 & 7.54 & 1000 & 234.13 & 18.74 & 1000 \\
3 & 1.60 & 22.50 & 200 & 80.66 & 20.58 & 200 \\
4 & 5.84 & 11.25 & 600 & 11.30 & 18.76 & 600 \\
5 & 0.17 & 18.75 & 600 & -13291.85 & 26.25 & 600 \\
6 & 3.53 & 27.52 & 1000 & 32.03 & 11.24 & 1000 \\
7 & -1.03 & 17.56 & 200 & -30.05 & 9.31 & 200 \\
8 & 0.71 & 18.75 & 200 & -323980.43 & 28.12 & 200 \\
9 & 2.65 & 7.50 & 1000 & 324721.68 & 26.25 & 1000 \\
10 & 2.01 & 2.52 & 200 & -391.01 & 16.81 & 200 \\
11 & 1.48 & 26.25 & 600 & -4.21 & 11.27 & 600 \\
12 & 1.20 & 26.25 & 1000 & -1241.57 & 3.75 & 1000 \\
13 & -629.95 & 3.75 & 200 & 1397.96 & 24.35 & 200 \\
14 & 629.51 & 3.75 & 600 & 20.57 & 3.76 & 600 \\
15 & -7.39 & 11.25 & 200 & 13018.81 & 1.86 & 200 \\
16 & 8.96 & 22.55 & 200 & 30.81 & 5.58 & 200 \\
\hline
\end{tabular}
\end{table}

\section{Simulation results and PV model parameters estimation}

\begin{figure*}[thbn]
\includegraphics[width=\textwidth]{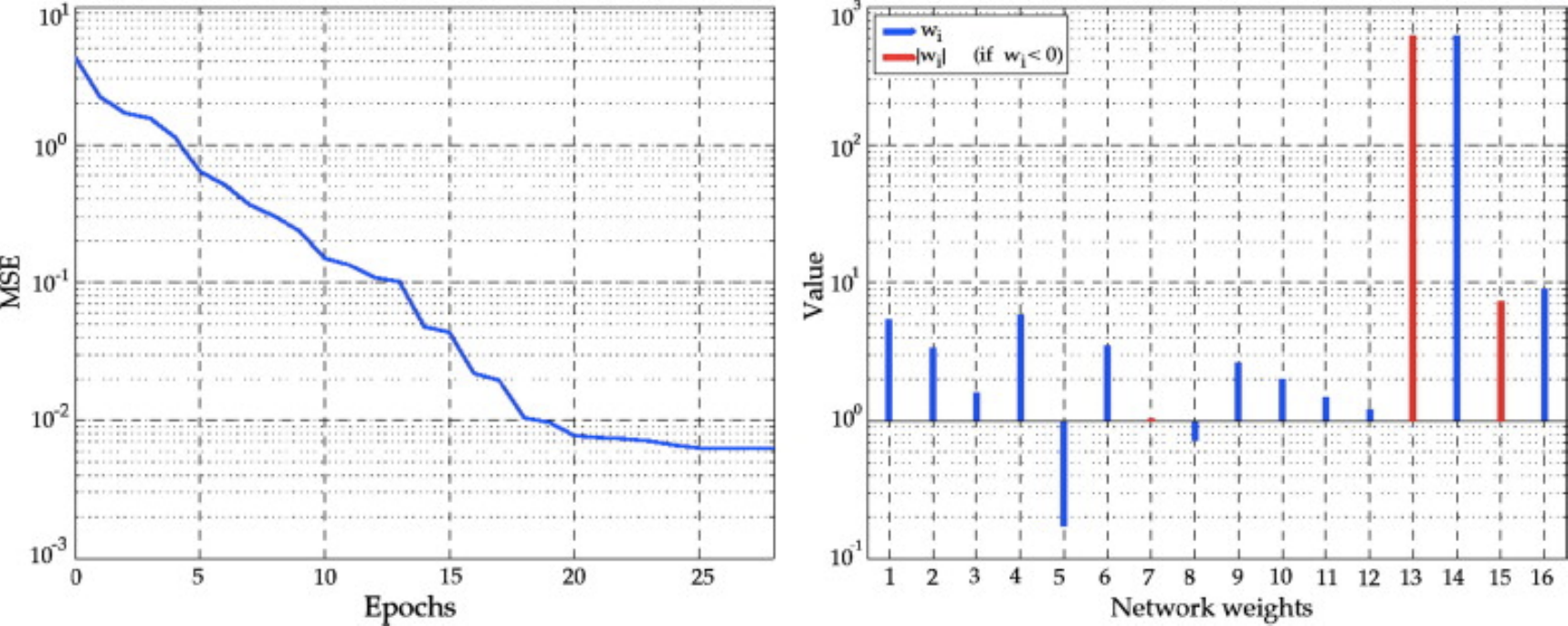}
\caption{Performance and weights for the I--V characteristics.}
\label{f3}
\end{figure*}

\begin{figure*}[thbn]
\includegraphics[width=\textwidth]{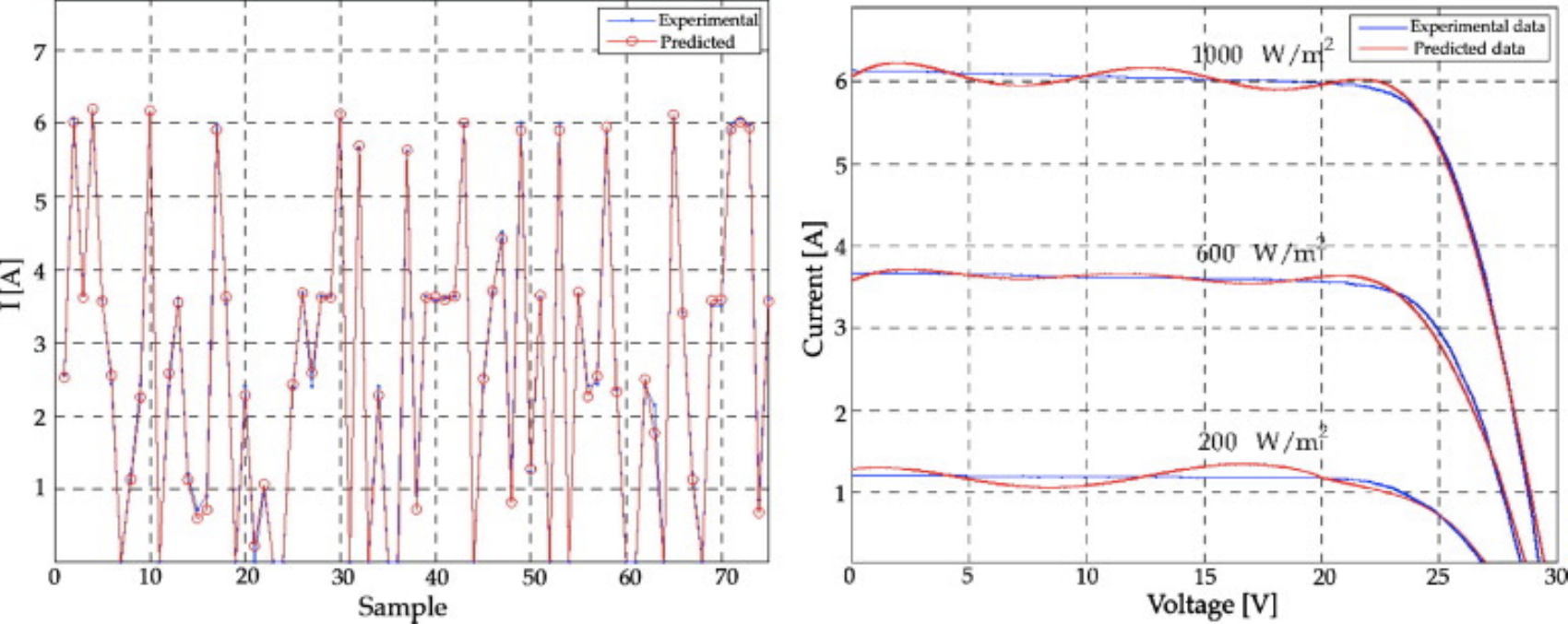}
\caption{ The I--V characteristics prediction by RBFNN based simulations.}
\label{f4}
\end{figure*}

\begin{figure*}[thbn]
\includegraphics[width=\textwidth]{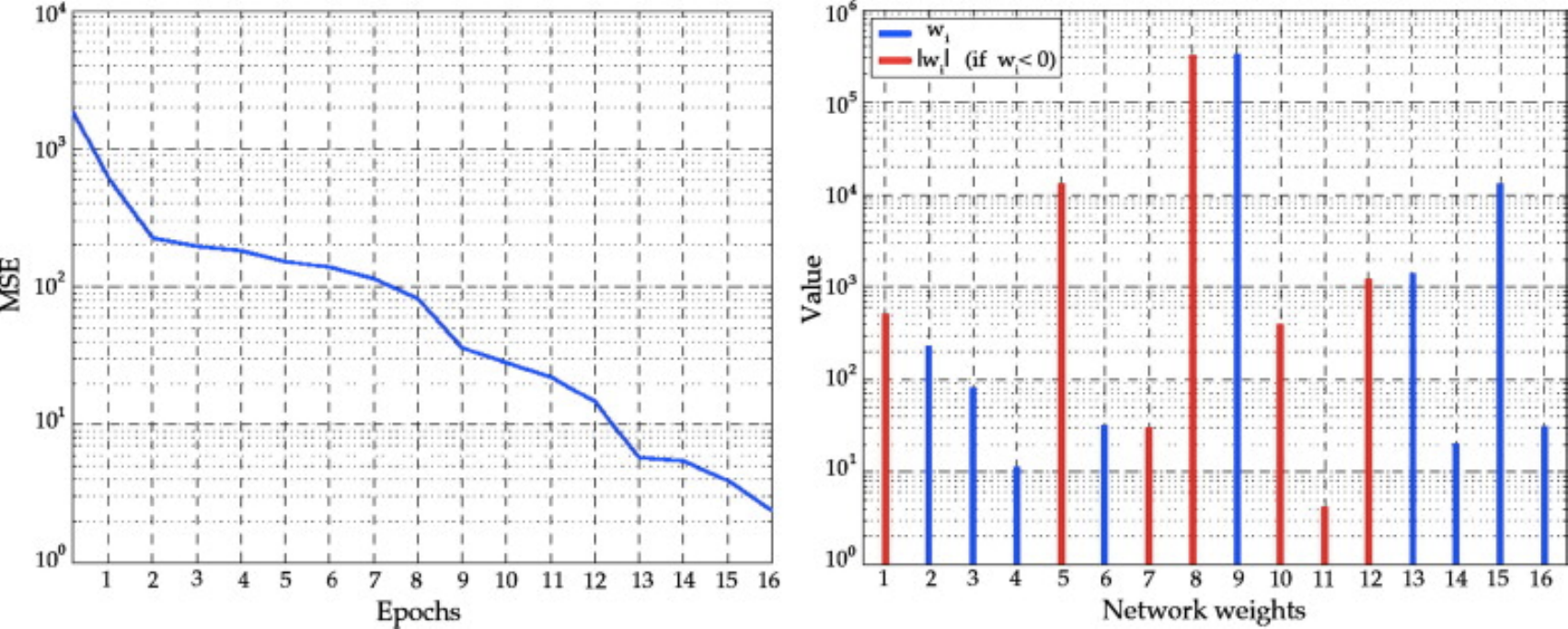}
\caption{Performance and weights for the P--V characteristics.}
\label{f5}
\end{figure*}

\begin{figure*}[thbn]
\includegraphics[width=\textwidth]{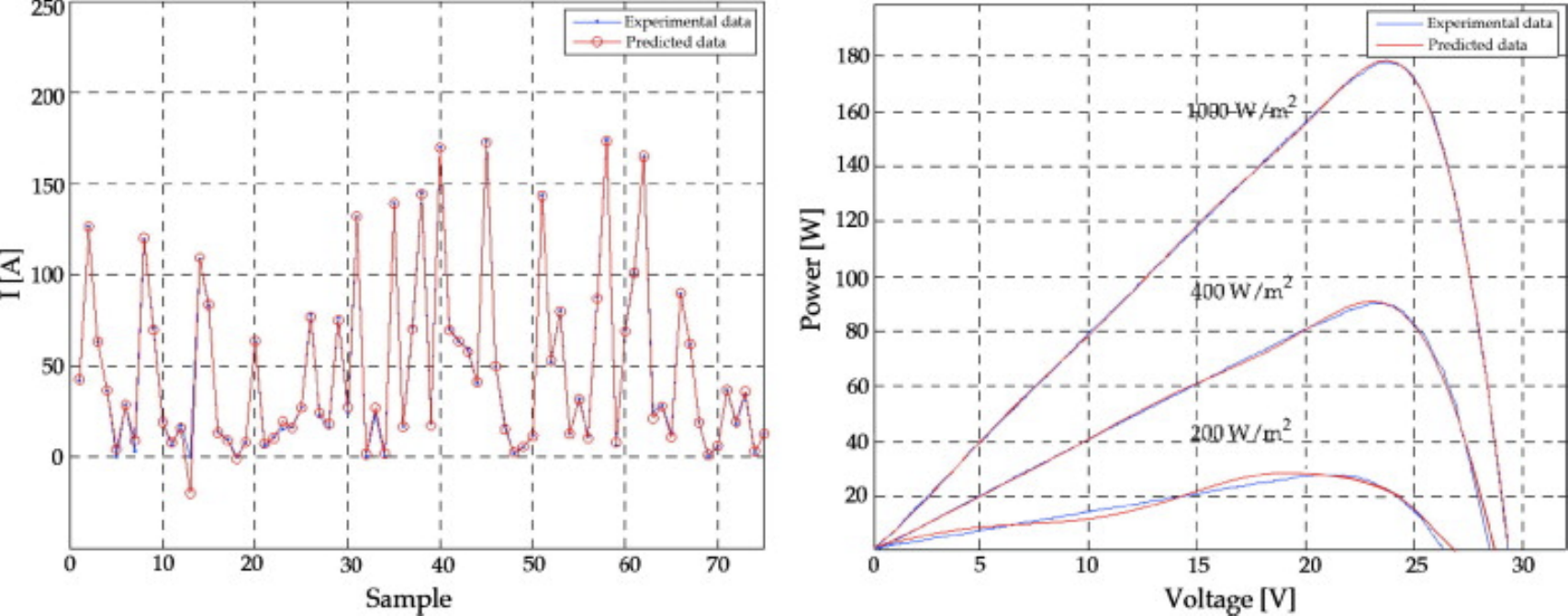}
\caption{The P--V characteristics prediction by RBFNN based simulations.}
\label{f6}
\end{figure*}

By using the expansion in power series we obtained a polynomial form for the generated current I of a PV module. By analysing the obtained polynomial form we make the implementation and calculation of the exponential $exp (−x^2)$ so finding the expression for the generated electric current by a summation until of 16 terms. The mathematical methods available in numerical analysis provide algorithms that utilize a numerical approximation for the problem of modeling the PV module components. Finally we find the model for a PV module by using the theory of non-linear circuits and systems. In Fig.~\ref{f3},Fig.~\ref{f4}, Fig.~\ref{f5} and Fig.~\ref{f6}, are reported the network’s performances and simulations results respectively for the I--V and P--V characteristics. It should be pointed out, that a larger spread width parameter for the RBFNN allows a smoother function approximation, but a too large spread width causes numerical problems, then the network training algorithm tries to reach the best trade-off between smoothness of approximation and numerical stability, that results the typical oscillations in RBFs approximation.

\section{Conclusion and discussion}

In this paper we proposed a novel neural network based technique for the modeling of solar cell output electrical characteristics. Using radial basis function neural network (RBFNN), we provide a model of the solar cell electrical behavior and the prediction of the I--V and P--V characteristics. By extensive computer simulations we gained that the RBFNN based models should achieve superior performance than the conventional neural models as multilayer perceptron (MLP). In literature the results enclosed in the paper are sometimes only at fixed solar irradiance. Neural models can be experienced at different solar radiation values. It’s common practice in industrial and scientific community to consider the shape of current and power versus the terminal voltage of a commercial PV. We propose a valuable RBFNN model that provide the shape of current and power as function of terminal voltage of the corresponding circuital electrical model at different value of G. Especially, for the estimation of I--V characteristics, the RBFNN based models match perfectly to the experimental characteristics at high value of solar irradiance superior precision is obtained as shown in the simulation results. By reviewing the elctrical I--V curves obtained by different models mainly based on electric circuits we point out that the five and seven parameters based model provide a better I--V and P--V estimation of the first part of curves \cite{a10}, while in the last part of the electrical output characteristics show a failure in the calculated data. Reversely the proposed RBFNN based model lead to an improvement along the final I--V and P--V curves then the electrical output behavior is now well evaluated in the most critical part. The numerical values of the computed I--V and P--V characteristics match closely to those obtained from the experimental data. Due to their low computational complexity, the RBFNNs are now preferred in modeling of complex and non-linear phenomena as solar cell characteristics. Furthermore the RBFNN can be used for other modeling purposes of solar cells as the five or the seven circuital parameters estimation.

\appendices
\section*{Acknowledgment}

This paper has been published in the final and reviewed version on  {\bf Applied Energy, vol. 97, pp. 956-961, 2012} \cite{aa}.

\bibliographystyle{IEEEtran}
\bibliography{napoliOA12apen1}
\end{document}